\documentclass[10pt,twocolumn,letterpaper]{article}

\usepackage{cvpr}
\usepackage{multirow}
\usepackage{bbding}
\usepackage[dvipsnames]{xcolor}

\definecolor{cvprblue}{rgb}{0.21,0.49,0.74}
\usepackage[pagebackref,breaklinks,colorlinks,citecolor=cvprblue]{hyperref}

\def\paperID{14081} 
\def\confName{CVPR}
\def\confYear{2024}

\title{MLIP: Enhancing Medical Visual Representation with Divergence Encoder and Knowledge-guided Contrastive Learning}

\author{Zhe Li \textsuperscript{1}, Laurence T. Yang \textsuperscript{1,2, ${*}$}, Bocheng Ren \textsuperscript{1}, Xin Nie \textsuperscript{1}, Zhangyang Gao \textsuperscript{3}, Cheng Tan \textsuperscript{3},  Stan Z. Li \textsuperscript{3}\\
	\textsuperscript{1} Huazhong University of Science and Technology \\
	\textsuperscript{2} Hainan University \\
	\textsuperscript{3} AI Lab, Research Center for Industries of the Future, Westlake University\\
	\thanks{$^{*}$Corresponding Author.}
	{\tt\small keycharon0122@gmail.com,\
		ltyang@ieee.org, \ bc.Revincent@gmail.com, \ niexin@hust.edu.cn,}
		\\ {\tt\small \{gaozhangyang,tancheng,stan.zq.li\}@westlake.edu.cn}}
\begin{document}
\maketitle
\begin{abstract}
The scarcity of annotated data has sparked significant interest in unsupervised pre-training methods that leverage medical reports as auxiliary signals for medical visual representation learning. However, existing research overlooks the multi-granularity nature of medical visual representation and lacks suitable contrastive learning techniques to improve the models' generalizability across different granularities, leading to the underutilization of image-text information. To address this, we propose MLIP, a novel framework leveraging domain-specific medical knowledge as guiding signals to integrate language information into the visual domain through image-text contrastive learning. Our model includes global contrastive learning with our designed divergence encoder, local token-knowledge-patch alignment contrastive learning, and knowledge-guided category-level contrastive learning with expert knowledge. Experimental evaluations reveal the efficacy of our model in enhancing transfer performance for tasks such as image classification, object detection, and semantic segmentation. Notably, MLIP surpasses state-of-the-art methods even with limited annotated data, highlighting the potential of multimodal pre-training in advancing medical representation learning.
\end{abstract}    
\section{Introduction}
Representation learning for medical radiographs has gained significant attention recently, owing to the availability of abundant annotated data. Numerous approaches \cite{esteva2017dermatologist, ronneberger2015u, de2018clinically, rajpurkar2018deep} have employed deep learning in a supervised manner to learn representations for downstream tasks. However, the acquisition of large-scale annotated data is time-consuming and costly. unsupervised pre-training methods have emerged as a promising alternative. These methods, which do not rely on annotated data, harness medical reports as ancillary signals that provide targeted supervision for visual representation learning. By incorporating language information, these models can acquire more universal visual representations that are transferable to downstream tasks and capable of domain transfer.

\begin{figure}
	\centering 
	\includegraphics[scale=0.35]{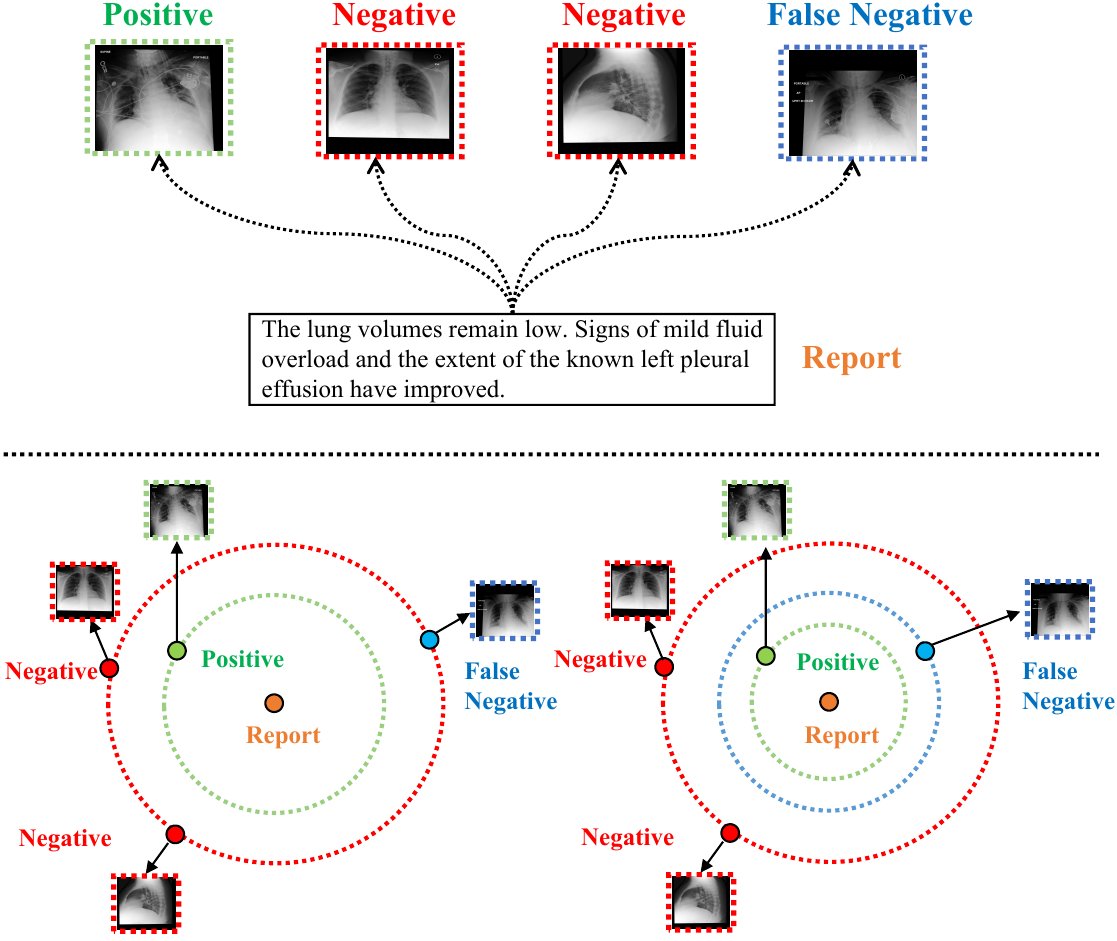} 
	\caption{Detailed illustration of false negatives in medical image-text. Conventional approaches consider false negative samples as negatives that are distant from positive samples in the lower left corner. In contrast, in the lower right corner, our proposed method distinguishes false negatives from negatives, effectively bringing them closer to positives.}  
	\label{motivation} 
\end{figure}
There are three mainstream paradigms in visual representation learning. Masked image modeling \cite{ he2022masked, wei2022masked} follows \textit{\textbf{mask-and-predict}} paradigm, randomly masking some patches and predicting missing information. Multimodal contrastive learning \cite{zhang2017tandemnet, hsu2018unsupervised, chauhan2020joint, zhang2022contrastive} conducts \textit{\textbf{embed-and-compare}} proxy tasks to maximize the mutual information between medical images and reports through image-text contrastive learning. Multi-view self-supervised learning \cite{caron2021emerging,he2020momentum,chen2020simple,chen2021exploring} adopts an \textit{\textbf{augment-and-compare}} paradigm, where an input image is randomly transformed into two augmented views and compare the two distinct views in the representation space. 

However, the fact that pathological features only occupy a small part of a radiograph means that a significant portion of the information may not be relevant for our analysis, decreasing the utilization of medical image-text data. Moreover, due to the unique nature of medical image-report compared to general text-image pairs, different symptoms may correspond to the same disease, and traditional contrastive learning will mistake samples that are not in the same batch as negative samples even if they are very close in the semantic space. In Fig \ref{motivation}, we purpose to differentiate between false negative and negative samples and further reduce the distance between false negative and positive samples.

Driven by the revelation from \cite{huang2021gloria, wangmulti, li2023vipmm}, we design a knowledge-guided \textit{\textbf{align-and-compare}} framework to capture multi-grained semantic information and to accurately align each image's pathology with the corresponding medical term \cite{huang2021gloria, lee2018stacked, li2019visual}. We introduce a knowledge-guided medical multimodal pre-trained model, dubbed MLIP, to explore the inherent multi-granularity cross-modal correspondence for enhancing the generalizability of visual representation. Specifically, we employ a combination of three distinct image-text contrastive learning methods to embed language into vision at different granularity and utilize two proxy tasks to establish the match between vision and language. Our model exploits multi-level correspondences between medical radiographs and reports to enhance generalized medical visual representation with contrastive learning. Our approach demonstrates state-of-the-art performance in image classification, object detection, and semantic segmentation, even when working with limited annotated data.

The key contributions are summarized as follows:
\begin{itemize}
	\item We introduce two dynamically updated \textbf{divergence encoders} for data augmentation, aiming to increase the number of samples and thus enhance the generalization ability of the model.
	\item  We propose to leverage cross-modal attention-based \textbf{token-knowledge-patch} alignment and incorporate contrastive learning to facilitate the exploration of local representations.
	\item We propose a \textbf{knowledge-guided prototype clustering} contrastive learning approach, which focuses on conducting contrastive learning at the category level rather than the individual samples.
	\item We pre-train MLIP on the MIMIC-CXR dataset \cite{johnson2019mimic}, evaluating the learned representations on seven downstream datasets. Experimental results demonstrate the superiority of our model over state-of-the-art methods, even with 1$\%$ and 10$\%$ training data.
\end{itemize}

\section{Related Work}
\subsection{Text-guided Medical Visual Representations Learning}

Medical reports are pivotal in unsupervised medical visual representation learning, with two primary methods dominating the field. The first method involves extracting disease labels from radiology reports using manually designed rules \cite{johnson2019mimic, irvin2019chexpert}, followed by pre-training image models for downstream tasks. However, defining the rules requires considerable human effort and domain expertise. On the other hand, the second method adopts image-text contrastive learning methods to integrate text and vision in an unsupervised manner \cite{de2018clinically, zhang2022contrastive, hsu2018unsupervised, huang2021gloria, wangmulti}. These methods have been shown remarkable performance in diverse downstream tasks, including medical object detection \cite{baumgartner2021nndetection}, image classification \cite{huang2021gloria, zhang2022contrastive}, and semantic segmentation \cite{zhang2022contrastive}. However, they have not effectively explored visual representations at different granularities and rely on partial semantic information.

To address these limitations, MGCA \cite{wangmulti} proposes to leverage multiple visual features at different granularities during the pre-training phase, enhancing the performance of models in downstream tasks. However, it overlooks the challenging sample issue in medical radiology. In this work, we propose a divergence encoder that manually updates its parameters based on the similarity between the output features and those of a common encoder. By increasing divergence between the two encoders, we enhance feature diversity and train the model to discriminate among similar samples effectively.
\begin{figure*}
	\centering 
	\includegraphics[width=0.94\textwidth]{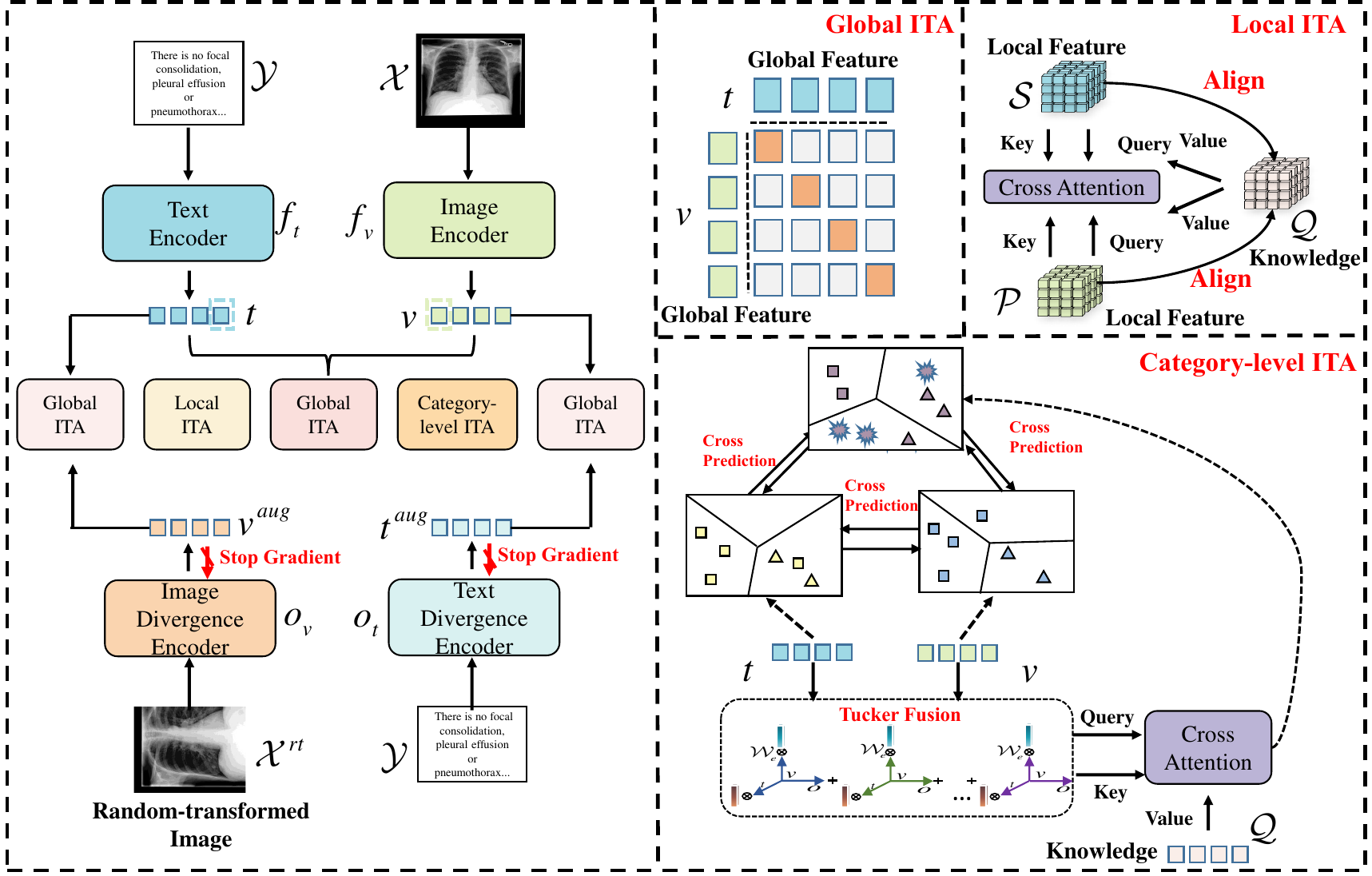} 
	\caption{Our model architecture employs global, local, and category-level image-text contrastive learning. Given medical images and reports as inputs, we extract global features and local features for each modality using image and text encoders. We leverage global features for global image-text contrastive learning, while the local features are aligned with domain-specific knowledge from UMLS to achieve fine-grained image-text alignment. Through tucker fusion and cross-modal attention mechanisms, we combine the image, text, and knowledge representations, facilitating category-level prototype contrastive learning. Furthermore, to enhance feature diversity, we introduce a divergence encoder as a data augmentation strategy, generating similar yet distinct features. This enables global contrastive learning between images and augmented text, as well as between text and augmented images.}  
	\label{frame} 
\end{figure*}
\subsection{Knowledge-guided Pre-training}
To enhance the model's knowledge and understanding ability by leveraging a broader background, numerous vision-and-language pre-training methods have been devised to incorporate domain-specific knowledge. These methods can be categorized into four distinct knowledge-guided schemes: embedding combination \cite{zhang2019ernie}, data structure compatibility \cite{he2020bert, liu2020k}, knowledge supervision \cite{wang2021kepler}, and neural-symbolic methods \cite{amizadeh2020neuro}. For instance, ERNIE-ViL \cite{yu2021ernie} introduces a vision and language alignment technique by utilizing a scene graph extracted from the input text. Similarly, KB-VLP \cite{chen2021kb} incorporates object tags from images and knowledge graph embeddings from texts to enhance the acquisition of knowledge-aware representations. ARL \cite{chen2022align} utilizes expert knowledge as an intermediate medium to align images and reports. Additionally, a recent study \cite{qin2022medical} proposes the automatic generation of visual and textual prompts, injecting expert medical knowledge into the prompt for pre-training.

In contrast to existing works, we propose an alignment method that leverages domain-specific knowledge as an intermediate mediator for aligning texts and images, along with a knowledge-guided prototype clustering contrastive learning. This approach integrates expert domain knowledge derived from the Unified Medical Language System (UMLS) \cite{bodenreider2004unified}. By incorporating UMLS knowledge into both vision and language modalities, our approach leverages knowledge as a medium to achieve improved alignment between images and text, facilitating more effective clustering of image-text pairs. Importantly, our method effectively mitigates the influence of disease-level false negatives without relying on object detectors or scene graph parsers.

\section{Proposed Approach}
In this section, we present our approach for learning effective medical visual representations using medical reports. We utilize a knowledge-guided \textit{align-and-compare} scheme, as depicted in Figure \ref{frame}, to match and align modalities and compare them in the representation space. Our method comprises four key components: 1) global image-text contrastive learning; 2) local token-knowledge-patch alignment contrastive learning; 3) knowledge-guided category-level contrastive learning; and 4) proxy tasks to ensure matching and prevent shortcut exploitation by the network. We discuss each component in detail in the following subsections and provide an overview of the overall training objective.

\subsection{Problem Setup}
Recently, it has been demonstrated in \cite{wangmulti,huang2021gloria} that learning medical visual representation learning without labels can achieve competitive performance. In this study, we follow the setting in \cite{wangmulti}, given a training set of $N$ medical image-report pairs $\mathcal D=\{(x_i, y_i)\}_{i=1, ..., N}$, we use an image encoder $f_v$ and a text encoder $f_t$ encode $\mathcal D$ to a global feature set $\mathcal E_{il}\!=\!\{(v_i, t_i)\!\!\!\mid \!\!\!v_i=f_v(x_i),t_i=f_t(y_i)\}_{i=1, ..., N}$, and a local feature set $\mathcal E_{tl}\!=\!\{(\mathcal P_i, \mathcal S_i)\}_{i=1, ..., N}$, where $\mathcal S_i=\{s^1_i, s^2_i, ..., s^V_i\} \in \mathbb{R}^{V\times d}$ and $\mathcal P_i=\{p^1_i, p^2_i, ..., p^{M^2}_i\}\in \mathbb{R}^{M^2\times d}$. $V$ denotes the length of the sentence and $M^2$ denotes the number of image patches. 

Furthermore, we incorporate expert knowledge into our model by constructing an extracted knowledge graph, as described in \cite{chen2022align}. This knowledge graph is denoted as $\mathcal{G}=\{(he_i, re_i, ta_i)\}_{i=1}^{N_{\mathcal{G}}}$, where $N_{\mathcal{G}}$ represents the number of graph triples, and $he_i$, $re_i$, and $ta_i$ correspond to the head entity, relation, and tail entity, respectively. The inclusion of this expert knowledge enhances the model's understanding and reasoning capabilities, enabling more informed alignment and representation learning.

\subsection{Global Image-text Contrastive Learning}
To pull correct samples closer and push random samples apart in the latent space, we follow \cite{tian2020contrastive, hjelm2018learning}, present a comprehensive discussion on global image-text contrastive learning by maximizing mutual information $\mathcal I(X,Y)$ between the vision element $X$ and the language component $Y$:
\begin{equation}\label{1}
	{\mathcal I(X,Y)\!=\!{\sum\limits}_{y\in Y}{\sum\limits}_{x\in X}P(x,y)\log \frac {P(x|y)}{P(x)}}.
\end{equation}

Eq.\ref{1} suggests that the fraction $\frac {P(x|y)}{P(x)}$ collapses to zero when $x$ and $y$ are incompatible with each other. Therefore, we hypothesize that $\frac {P(x|y)}{P(x)}$ is proportional to the similarity between $x$ and $y$. Further, the maximization of mutual information corresponds to the maximization of the similarity $\textrm{sim}(x, y)$ between $x$ and $y$, which can be represented as:
\begin{equation}\label{2}
	\mathcal I(v,t)\propto \mathcal I(X,Y)\propto \textrm{sim}(x, y)\propto \textrm{sim}(v,t).
\end{equation}

Specifically, inspired by \cite{chen2020simple}, we firstly utilize two projection layers $h_v$ and $h_t$ to map $v_i$ and $t_i$ into a normalized shared feature space, yielding $v_i^{\ast} \in \mathbb R^d$ and $t_i^{\ast} \in \mathbb R^d$, respectively. Then, we apply the dot product to model the similarity between $v_i^{\ast}$ and $t_i^{\ast}$. To obtain more effective features, we perform Self-Attention \cite{vaswani2017attention} and LayerNorm \cite{ba2016layer} on features:
\begin{subequations}
	\begin{align}
		&v_i^{\ast}=\text{LN}(\text{SA}\{h_v(v_i)\});\\
		&t_i^{\ast}=\text{LN}(\text{SA}\{h_t(t_i)\}),\\
		&\textrm{sim}\{v_i^{\ast}, t_i^{\ast}\}=v_i^{\ast}{t_i}^T,
	\end{align}
\end{subequations}
where $\text{SA}$ denotes Self-Attention module and $\text{LN}$ denotes LayerNorm module.

We optimize this process via image-text contrastive loss based on InfoNCE loss \cite{van2018representation}, which are designed to maximize the mutual information between the correct image-text pairs in the latent space:
\begin{subequations}
	\begin{align}
		\mathcal L^{il}_{v2t}(v_i,t_i)=-\log(
		\frac{\phi_{il}(v_i,t_i)}{{\sum\limits}_{k=1}^B\phi_{il}(v_i, t_k)}), \label{3}
		\\
		\mathcal L^{il}_{t2v}(v_i,t_i)=-\log(
		\frac{\phi_{il}(v_i,t_i)}{{\sum\limits}_{k=1}^B\phi_{il}(v_k, t_i)}),\label{5}
	\end{align}
\end{subequations}
where $\phi_{il}(v_i,t_i)=\exp(\frac{\textrm{sim}(v_i^{\ast}, t_i^{\ast})}{\tau_1})$, $B$ is the batch size and $\tau_1$ is the global temperature hyper-parameter.

Directly optimizing $\mathcal I(v, t)$ is a challenging task. As an alternative,  \cite{van2018representation} has proposed an alternative method to optimize the lower bound of mutual information:
\begin{equation}\label{alt}
	{\mathcal I(v,t)}\geq \log N^{'}-\mathcal L^\text{NCE}(v,t),
\end{equation}
where $N^{'}$ is the number of negative samples. In Eq.\ref{alt}, minimizing $\mathcal L^{\text{NCE}}(v,t)$ is equivalent to maximizing the lower bound of the mutual information between the medical image and the corresponding report. 

To increase the number of samples and enhance the feature diversity, we perform a divergence encoder to achieve data augmentation and extend the gap between samples. We define image divergence encoder $o_v$ and text divergence encoder $o_t$, initialized by $f_v$ and $f_t$, respectively. Then we obtain features incrementally differentiated from $v_i$ and $t_i$:
\begin{equation}\label{momentum}
	v_i^{aug}=o_v(x^{rt}_i);t_i^{aug}=o_t(y_i),
\end{equation}
where $x^{rt}_i$ denotes randomly transformed images. We manually update divergence encoders' parameters instead of relying on backpropagation:
\begin{subequations}\label{param}
	\begin{align}
		&\theta_{o_t}\!\!=\!\!s_t\!*\!\theta_{f_t} + (1-s_t)\!*\!\theta_{o_t},
		\\
		&\theta_{o_v}\!\!=\!\!s_v\!*\!\theta_{f_v} + (1-s_v)\!*\!\theta_{o_v},
	\end{align}
\end{subequations}
where $s_t=\text{cosine}(t_i,t_i^{aug})$ and $s_v=\text{cosine}(v_i,v_i^{aug})$, and $\theta_{o_t}, \theta_{o_v}, \theta_{f_t}, \theta_{f_v}$ are the parameters of $o_t, o_v, f_t, f_v$, respectively. In this way, as the $s_v (s_t)$ increases, we aim to retain fewer parameters from $o_v (o_t)$ and incorporate more parameters from $f_v (f_t)$, in order to generate more diverse features. Then we use Eq.\ref{3}, \ref{5} to compute $\mathcal L_{v2a}^{il}$ and $\mathcal L_{avt}^{il}$.

We compute the objective $\mathcal L_{ita}$ as the average of the four loss values:
\begin{equation}\label{7}
	\begin{split}
		\mathcal L_{ita} = \frac{1}{2N}{\sum\limits}^{N}_{i=1}(\mathcal L_{v2t}^{il}(v_i,t_i)+\mathcal L_{t2v}^{il}(v_i,t_i))\\
		+\frac{\lambda_0}{2N}{\sum\limits}^{N}_{i=1}(\mathcal L_{v2a}^{il}(v_i,t_i^{aug})+\mathcal L_{avt}^{il}(v_i^{aug},t_i)),
	\end{split}
\end{equation}
where $N$ is the total number of samples and $\lambda_0$ denotes the weight for augmented image-text contrastive learning.
\subsection{Local Token-knowledge-patch Alignment Contrastive Learning}
In medical images, pathologies are often visually subtle and occupy a small fraction of the overall image, while only a few disease-related tags in the associated report accurately depict the critical medical condition. Given this observation, we employ a local image-text contrastive learning method to maximize the mutual information between local features and achieve cross-modal alignment between images and texts, inspired by \cite{wangmulti, cui2020unsupervised}.

However, traditional token-patch alignment contrastive learning is utilizing the local features of the image and text to compute the attention matrix, and then perform contrastive learning after aligning the images and texts. Since medical radiology is highly professional and there is a certain bias between different datasets, we regard professional knowledge from the UMLS \cite{bodenreider2004unified} as a medium between vision and language. To achieve more accurate token-patch alignment, we align the knowledge with radiographs and reports.

Similar to global feature, we apply Self-Attention and LayerNorm module on every features:
\begin{equation}
	p_i=\text{LN}(\text{SA}\{h_v(p_i)\}); s_i=\text{LN}(\text{SA}\{h_t(s_i)\}).
\end{equation}

We apply the knowledge representation learning algorithm TransE \cite{bordes2013translating} to the knowledge graph $\mathcal{G}$ to obtain entity embeddings. Subsequently, we utilize the Graph Attention Network \cite{velickovic2017graph} to capture local information in the graph neighborhood for each node. This allows us to obtain knowledge representations, denoted as $\{e_i\}_{i=1}^{N_e} \in \mathbb{R}^{N_e\times d_e}$, where $d_e$ represents the feature dimension and $N_e$ denotes the number of entity.

We adopt cross-modal attention mechanism \cite{chen2020uniter, lu2016hierarchical} to explore the matching between knowledge and image:
\begin{subequations}
	\begin{align}
		attn_{j,k}&^{vk}=\text{softmax}(\frac{(Qp_i^j)^T(Ke_i^k)}{\sqrt{d}}),
		\\
		{zv}_i^j&={\sum\limits}_{k=1}^Nattn_{j,k}^{vk}(Ve_i^k),
	\end{align}
\end{subequations}
where $Q, K, V \in \mathbb{R}^{d\times d}$ are trainable matrices. $e_i$ is mapped to $\mathbb{R}^{M^2\times d}$. ${zv}_i^j$ is cross-modal knowledge embedding corresponding to $p_i^j$.

Lying in the purpose of maximizing the lower bound of mutual information, we leverage InfoNCE loss \cite{van2018representation} to pull $p_i^j$ and ${zv}_i^j$ closer and push $p_i^j$ and other cross-modal knowledge embeddings apart. However, given that irrelevant information only occupies a vast majority of medical images, we employ $w_i^j$ to balance the weights of different patches. The loss $\mathcal L_{v2t}^{tl}$ is designed symmetrically as:
\begin{equation}\label{7}
	\begin{split}
		\mathcal L_{v2t}^{tl}=-\frac{1}{2NM^2}&\sum\limits_{i=1}^N\sum\limits_{j=1}^{M^2}w_i^j(\log\frac{\phi_{tl}({p_i^j},{{zv}_i^j})}{{\sum\limits}_{k=1}^{M^2}\phi_{tl}({p_i^j},{{zv}_i^k})}\\
		&+\log\frac{\phi_{tl}({{zv}_i^j}, {p_i^j})}{{\sum\limits}_{k=1}^{M^2}\phi_{tl}({{zv}_i^k}, {p_i^j})}),
	\end{split}
\end{equation}
where $\phi_{tl}(p_i^j,{zv}_i^j)=\exp(\frac{\textrm{sim}(p_i^j, {zv}_i^j)}{\tau_2})$, $\tau_2$ is the local temperature hyper-parameter. To establish the correlation between the $j$-th visual patch and the [CLS] token, we assign the weight $w_i^j$ using the last-layer attention mechanism averaged across multiple heads. 

Similarly, for the $j$-th text token, we calculate corresponding cross-modal knowledge embedding ${zt}_i^j$ and construct local contrastive loss $\mathcal L_{t2v}^{tl}$ to maximize the lower bound of mutual information between $s_i^j$ and ${zt}_i^j$. The objective $\mathcal L_{tl}$ can be defined as the average of these two losses:
\begin{equation}\label{8}
	\mathcal L_{tl} = \frac{1}{2}(\mathcal L_{v2t}^{tl} + \mathcal L_{t2v}^{tl}).
\end{equation}
\subsection{Knowledge-guided Category-level Contrastive Learning}

For a given radiograph-report pair, traditional contrastive learning approaches treat other radiograph-report pairs within the same batch as negative samples. However, in the context of category-level analysis, samples that belong to different batches but exhibit highly similar semantics should be considered positive samples. In our approach, we aim to select representative samples in each iteration, emphasizing their ability to capture meaningful disease-related information. In the medical domain, expert knowledge plays a crucial role in representation learning. We purpose to bridge the gap between the vast knowledge learned from general visual and textual data and its effective application in the intricate realm of medical radiology. Therefore, we incorporate expert knowledge from UMLS \cite{bodenreider2004unified} as an auxiliary signal.  Drawing inspiration from \cite{caron2020unsupervised, qin2022medical}, we propose a knowledge-guided clustering-based approach to improve the efficacy of learned representations. We bring together highly similar samples with high-level semantics, even when originating from different batches, and ensure their proximity in the feature space, rather than increasing their distance from one another.

Motivated by \cite{liu2021exploring}, we realize to filter out irrelevant information and explore more fine-grained relations between images and text. To achieve this, we employ a mechanism that identifies the most relevant topic in a given context. Specifically, we utilize $v_i^{\ast}$ to find the most relevant topic in $t_i^{\ast}$, resulting in $\dot{t_i}$. Then, we use $\dot{t_i}$ to find the relevant topic in $v_i^{\ast}$, leading to $\dot{v_i}$. The process is mathematically defined as follows:
\begin{equation}
	\dot{t_i}\!=\!\text{LN}(\text{softmax}(\frac{{v_i^{\ast}}^Tt_i^{\ast}}{\sqrt{d}})t_i^{\ast}); 		\dot{v_i}\!=\!\text{LN}(\text{softmax}(\frac{{v_i^{\ast}}^Tv_i^{\ast}}{\sqrt{d}})\dot{t_i}),
\end{equation}
then we utilize tucker fusion \cite{ben2017mutan} to seamlessly integrate visual and textual features, further fuse with knowledge representations:
\begin{equation}
	\mathcal Q=(({\mathcal T_c}\times_{1}{\dot{v_i}})\times_{2}{\dot{t_i}})\times_{3}\mathcal W_o,
\end{equation}
where $\mathcal W_o$ represents a mapping matrix which is trainable and maps fused features to a certain dimensional space, and $\mathcal T_c$ denotes the core tensor.

To further integrate knowledge with modality-specific features, we employ a linear mapping layer to project the knowledge representation $e_i$ into a $d$-dimensional space and incorporate it with fused features using cross-modal attention, thereby facilitating the fusion of information across modalities:
\begin{equation}
	vkt_i = \text{SA}(\text{softmax}(\frac{{\mathcal Q}^Te_i}{\tau_3})\cdot e_i),
\end{equation}
where $\tau_3$ is the temperature hyper-parameter we set to scale the attention.

For image-text features pair $(\dot{v_i}, \dot{t_i})$ and knowledge-fused features, we apply the iterative Sinkhorn-Knopp clustering algorithm \cite{cuturi2013sinkhorn} to generate a cluster assignment code $u^{vkt,i} \in \mathbb{R}^{C}$, by assigning $vkt_i$ to $C$ clusters separately. To facilitate this, we introduce a set $\mathcal{J} = {j_1, ..., j_C}$ that contains $C$ trainable cross-modal prototypes, where each prototype $j_c \in \mathbb{R}^d$. We calculate the visual softmax probability $p^{v,i}$ by computing the cosine similarity between the visual feature vector $\dot{v_i}$ and all cross-modal prototypes in $\mathcal{J}$. Similarly, the textual softmax probability $p^{t,i}$ is obtained by measuring the cosine similarity between the textual feature vector $\dot{t_i}$ and all cross-modal prototypes in $\mathcal{J}$:
\begin{equation}
	p^{v,i}_c=\frac{\exp({\dot{v_i}}^Tj_c/\tau_4)}{{\sum\limits}_l\exp({\dot{v_i}}^Tj_l/\tau_4)}; p^{t,i}_c=\frac{\exp({\dot{t_i}}^Tj_c/\tau_4)}{{\sum\limits}_l\exp({\dot{t_i}}^Tj_l/\tau_4)},
\end{equation}
where $\tau_4$ is a category-level temperature hyper-parameter and $c$ denotes the $c$-th element of the vector.

To enable knowledge-guided category-level contrastive learning, we employ $u^{vkt,i}$ as the pseudo-label for training $\dot{t_i}$ and $\dot{v_i}$. This allows the three features to interact in the latent space and guide the shifting of positive and negative samples with the assistance of domain-specific knowledge. The objective loss $\mathcal L_{cl}$ is formulated as follows:
\begin{equation}
	\mathcal L_{cl}\!=\!\frac{1}{2N}\sum\limits_{i=1}^N\sum\limits^C_{c=1}u^{vkt,i}_c\log p^{v,i}_c\!+ u^{vkt,i}_c\log p^{t,i}_c.
\end{equation}

\subsection{Image-text Matching and Text Swapping}
In order to identify the alignment between radiographs and their corresponding reports, we propose two pretext tasks aimed at bridging the semantic divide between visual and linguistic information within the feature space: 1) computing relevance scores between image patch and contextualized sentence to evaluate the degree of correlation between the image and text elements; 2) randomly substituting medical reports corresponding to the image with a predetermined probability, improving the discriminative ability on mismatched samples of the model. 

We assume that the text features $t$ and image features $v$ have been normalized. Therefore, we construct the similarity between the two modalities as a relevance score:
\begin{equation}
	r(v, t) = v^{T} \cdot t,
\end{equation}
subsequently, we randomly select another image $v^{'}$ and obtain its corresponding relevance
score $r(v^{'}, t)$. To ensure that the difference between $r(v, t)$ and
$r(v^{'}, t)$ is greater than a pre-specified margin $\mathcal G$, we utilize the
hinge loss function to compute image-text match loss:
\begin{equation}
	\mathcal L_{itm} = \max(0, \mathcal G-r(v, t)+r(v^{'}, t)).
\end{equation}

Similarly, we propose a text swapping task, which involves randomly replacing text with a predefined probability $\gamma$. We employ a bidirectional similarity Hinge loss to penalize the model for insufficient discriminative ability. This task aims to enhance the model's ability to distinguish between different reports. We employ a cross-modal attention mechanism to fuse the text and image modalities, then compute the relevance score by performing a weighted summation of the similarity between the fused representation and the original text-image pair. Our objective is to ensure that this score exceeds the score obtained after replacing the text by a margin $\mathcal G^{'}$:
\begin{subequations}
	\begin{align}
		r&_{ts}(v, t)=v^{T}\cdot t+\alpha\cdot \text{CA}(v, t)^{T}\cdot\text{CA}(t, v),
		\\
		r_{ts}&(v, t^{'})=v^{T}\cdot t+\alpha\cdot \text{CA}(v, t^{'})^{T}\cdot\text{CA}(t^{'}, v),
		\\
		&\mathcal L_{ts} = \max(0, \mathcal G{'}-r_{ts}(v, t)+r_{ts}(v^{'}, t)),
	\end{align}
\end{subequations}
where $\text{CA(x,y)}=\text{softmax}(\frac{x^{T}\cdot y}{\sqrt{d}})\cdot y$.
Through these two designed proxy tasks, we compute the image-text matching loss $\mathcal L_{itm}$ and the text swapping loss $\mathcal L_{ts}$. These losses quantify the model's ability to accurately match radiographs to their appropriate reports, thereby providing a measurable objective for the optimization process. 
\subsection{Overall Objective}
Our training approach involves joint optimization of the five losses, aiming to promote the acquisition of effective and generalizable medical image representations by the network. The overall training objective can be expressed as follows:
\begin{equation}
	\mathcal L\!=\!\lambda_1\mathcal L_{il}+\lambda_2\mathcal L_{tl}+\lambda_3\mathcal L_{cl}+\lambda_4\mathcal L_{itm} + \lambda_5\mathcal L_{ts},
\end{equation}
where $\lambda_1$, $\lambda_2$, $\lambda_3$, $\lambda_4$ and $\lambda_5$ are hyper-parameters employed to balance the weights associated with each respective loss. 

\section{Experiments}
\subsection{Pre-training Dataset and Implementation Details}
Our MLIP framework is initially pre-trained on the MIMIC-CXR 2.0.0 dataset \cite{johnson2019mimic}, with data consistency ensured through preprocessing methods from \cite{zhang2022contrastive}. Lateral views are excluded from the dataset as downstream datasets only include frontal-view chest images. Inspired by \cite{wangmulti}, we extract impression and finding sections from free-text reports, providing comprehensive descriptions of medical diseases. We filter out empty or short reports, resulting in approximately 217,000 image-text pairs. Details about our implementation can be found in the supplement \ref{sec:rationale}.
\subsection{Downstream Tasks}
\paragraph{Medical Object Detection.}
We assess the capability of our pre-trained image encoder for medical object detection on the \textbf{RSNA} Pneumonia dataset \cite{shih2019augmenting} (stage 2 version) and the \textbf{ Object CXR} dataset \cite{healthcare2020object}. The detection performance is evaluated using the YOLOv3 \cite{farhadi2018yolov3} frozen setting, where the pre-trained ResNet-50 \cite{he2016deep} image encoder acts as a fixed backbone for YOLOv3. In this configuration, only the classification layers are fine-tuned. To evaluate the efficiency of data utilization, we conduct experiments in the zero-shot scenario and further fine-tune the model using 1\%, 10\%, and 100\% of the available training data. Evaluation is performed using the Mean Average Precision (mAP) metric, computed with IOU thresholds ranging from 0.4 to 0.75.
\begin{table}[ht]
	\setlength{\belowdisplayskip}{-5cm}
	\centering
	\resizebox{\linewidth}{!}{
		\begin{tabular}{cccccccccccc}
			\toprule
			\multirow{2}*{Method} & \multicolumn{4}{c}{RSNA(mAP)} & \multicolumn{4}{c}{Object CXR(mAP)}  \\
			&Zero-shot& 1\% & 10\% & 100\% & Zero-shot&1\% & 10\% & 100\%
			\\ 
			\midrule
			Random Init &$\sim$& 1.0&4.0& 	8.9&$\sim$&$\sim$&$\sim$& 4.4 \\
			ImageNet Init &$\sim$& 3.6 &8.0 & 15.7& $\sim$&$\sim$& 8.6& 15.9 \\
			ConVIRT \cite{zhang2022contrastive}&3.7 &8.2& 15.6& 17.9&$\sim$& $\sim$&8.6&15.9\\
			GLoRIA-CheXpert \cite{huang2021gloria}&4.4 &9.8& 14.8& 18.8& $\sim$ &$\sim$&10.6 &15.6 \\
			GLoRIA-MIMIC \cite{huang2021gloria}&6.2& 10.3& 15.6&23.1& $\sim$&$\sim$& 8.9&16.6\\
			MGCA (ResNet-50) \cite{wangmulti}& \textcolor{blue}{7.8}& \textcolor{blue}{12.9}& \textcolor{blue}{16.8}&\textcolor{blue}{24.9} & $\sim$ & $\sim$&\textcolor{blue}{12.1}& \textcolor{blue}{19.2}\\
			MLIP (Ours, ResNet-50)&  \textcolor{red}{12.3}& \textcolor{red}{17.2}& \textcolor{red}{19.1}& \textcolor{red}{25.8}&\textcolor{red}{2.7}& \textcolor{red}{4.6}&\textcolor{red}{17.4} & \textcolor{red}{20.2}\\
			\bottomrule
	\end{tabular}}
	\caption{Fine-tuned results (mAP [\%]) of object detection with 1\%, 10\%, and 100\% of the available training data in RSNA and Object CXR. $\sim$ means mAP is smaller than 1\%.}
	\label{od}
\end{table}
\paragraph{Medical Semantic Segmentation.}
We evaluate the performance of our model for medical semantic segmentation on the \textbf{SIIM} Pneumothorax dataset \cite{siim-acr-pneumothorax-segmentation} and the \textbf{RSNA} Pneumonia dataset \cite{shih2019augmenting}. Following the methodology presented in \cite{huang2021gloria}, we adopt the fine-tuning protocol of U-Net \cite{ronneberger2015u} to assess the segmentation task. Specifically, we utilize the pre-trained ResNet-50 image encoder as a fixed backbone for the U-Net architecture and train the decoder component using varying proportions of the available training data (1\%, 10\%, and 100\%). We also evaluate our model in the zero-shot scenario. To evaluate the quality of segmentation, we compute Dice scores \cite{wang2020image} as the chosen metric for performance assessment.

\begin{table}[h]
	\setlength{\abovedisplayskip}{-0.5cm}
	\centering
	\resizebox{\linewidth}{!}{
		\begin{tabular}{cccccccccccc}
			\toprule
			\multirow{2}*{Method} & \multicolumn{4}{c}{RSNA(Dice)} & \multicolumn{4}{c}{SIIM(Dice)}  \\
			& Zero-shot&1\% & 10\% & 100\% &Zero-shot& 1\% & 10\% & 100\%
			\\ 
			\midrule
			Random Init & 3.9&6.9 & 10.6 & 	18.5 & $\sim$&9.0 & 28.6 & 54.3 \\
			ImageNet Init & 17.6&34.8 &39.9 & 64.0 & 2.2&10.2 & 35.5 & 63.5 \\
			ConVIRT \cite{zhang2022contrastive}& 23.3&55.0& 67.4& 67.5& 11.7&25.0&43.2 &59.9\\
			GLoRIA-CheXpert \cite{huang2021gloria}& 32.0&59.3& 67.5& 67.8& 19.8&35.8 &46.9 & 63.4 \\
			GLoRIA-MIMIC \cite{huang2021gloria}& 34.6&60.8& 68.2& 67.6& 21.0&37.6& 56.4&64.0\\
			MGCA (ResNet-50) \cite{wangmulti}&  \textcolor{blue}{34.9}&\textcolor{blue}{63.0}& \textcolor{blue}{68.3}&\textcolor{blue}{69.8} & \textcolor{blue}{33.5}&\textcolor{blue}{49.7} & \textcolor{blue}{59.3}& \textcolor{blue}{64.2}\\
			MLIP (Ours, ResNet-50)&\textcolor{red}{44.3}&  \textcolor{red}{67.7}& \textcolor{red}{68.8}& \textcolor{red}{73.5}&\textcolor{red}{40.2}& \textcolor{red}{51.6} & \textcolor{red}{60.8}&\textcolor{red}{68.1}\\
			\bottomrule
	\end{tabular}}
	\caption{We present the semantic segmentation results (Dice [\%]) achieved on the SIIM and RSNA datasets. Each dataset is fine-tuned using 1\%, 10\%, and 100\% of the available training data. The best results obtained for each setting are highlighted in red, while the suboptimal results are highlighted in blue.}
	\label{Table 3}
\end{table}

\paragraph{Medical Image Classification.}
We perform medical image classification on the \textbf{RSNA} Pneumonia dataset \cite{shih2019augmenting}, \textbf{COVIDx} dataset \cite{wang2020covid}, and \textbf{CheXpert} dataset \cite{irvin2019chexpert}. To evaluate the transferability of our pre-trained image encoder, we adopt the Linear Classification setting following the methodology proposed in prior work \cite{huang2021gloria, wangmulti}. This involves freezing the pre-trained ViT-B/16 \cite{dosovitskiy2020image} or ResNet-50 image encoder and training only a linear classification head, initialized randomly, for the downstream classification task. Additionally, to assess data efficiency, we conduct experiments in the zero-shot scenario and evaluate the model using 1\%, 10\%, and 100\% of the training data for each classification dataset. The evaluation metrics used are the area under the receiver operating characteristic (ROC) curve (AUROC) for RSNA and CheXpert, and accuracy (ACC) for COVIDx-v6, consistent with the evaluation criteria outlined in \cite{zhang2022contrastive}. More details can be found in the supplementary \ref{sec:Implementation of Downstream tasks}.

\begin{table*}\huge
	\setlength{\abovedisplayskip}{-0.0cm}
	\centering
	\resizebox{\linewidth}{!}{
		\begin{tabular}{cccccccccccccccc}
			\toprule
			\multirow{2}*{Method} & \multicolumn{4}{c}{CheXpert(AUC)} & \multicolumn{4}{c}{RSNA(AUC)} & \multicolumn{4}{c}{COVIDx(ACC)}  \\
			& Zero-shot & 1\% & 10\% & 100\% & Zero-shot & 1\% & 10\% & 100\% & Zero-shot & 1\% & 10\% & 100\%
			\\ 
			\midrule
			Random Init & - & 56.1 & 62.6 & 	65.7 & - & 58.9 & 69.4 & 	74.1 & - & 50.5 & 60.3 & 70.0 \\
			ImageNet Init & - & 74.4 & 79.7 & 	81.4& - & 74.9 & 74.5 & 76.3 & - & 64.8 & 78.8 & 86.3 \\
			\midrule
			\textbf{pre-trained on CheXpert}\\
			DSVE \cite{engilberge2018finding}& 26.6& 50.1 & 51.0 & 	51.5& 18.7 &49.7& 52.1 & 57.8 & -& - & - & - \\
			VSE++ \cite{faghri2017vse++}& 27.3 & 50.3 & 51.2 & 	52.4& 19.1& 49.4 & 57.2 & 67.9& -& - & - & -\\
			GLoRIA \cite{huang2021gloria}& 50.4 & 86.6 & 87.8 & 	88.1& 39.2& 86.1 & 88.0 & 88.6 & 20.9 & 67.3 & 77.8 & 89.0\\
			\midrule
			\textbf{pre-trained on MIMIC-CXR}\\
			Caption-Transformer \cite{cornia2020meshed} &42.2&77.2& 82.6 & 83.9  & - & -& - & -& - & -& - & -\\
			Caption-LSTM \cite{xu2015show}& 45.6& 85.2 & 85.3 & 86.2 & - & -& - & -& - & -& - & -\\
			Contrastive-Binary \cite{tan2019lxmert}& 46.8&84.5 & 85.6 & 85.8 & - & -& - & -& - & -& - & -\\
			ConVIRT \cite{zhang2022contrastive}& 47.6 & 85.9 & 86.8 & 	87.3& 34.7& 77.4 & 80.1 & 81.3 & 17.8 & 72.5 & 82.5 & 92.0\\
			GLoRIA-MIMIC \cite{huang2021gloria}& 51.7 & 87.1 & 88.7 & 88.0& 40.6& 86.6 & 89.2 & 90.4& 22.1 & 67.3 & 81.5 & 88.6\\
			MGCA (ResNet-50) \cite{wangmulti}&50.2& 87.6 & 88.0 & 	88.2& 41.0& 88.6 & 89.1 & 89.9 & 24.5 & 72.0 & 83.5 & 90.5\\
			MGCA (ViT-B/16) \cite{wangmulti}& 50.0 & \textcolor{blue}{88.8} & \textcolor{blue}{89.1}& 	\textcolor{blue}{89.7}& 39.2& \textcolor{blue}{89.1} & \textcolor{blue}{89.9} & \textcolor{red}{90.8}& \textcolor{blue}{33.2} & 74.8& 84.8 & \textcolor{blue}{ 92.3}\\
			MLIP (Ours, ResNet-50)& \textcolor{blue}{56.9}  & 87.8 & 88.7 & 	88.9&  \textcolor{blue}{42.9}& 88.8 & 89.6 & \textcolor{blue}{90.6} & 26.3& \textcolor{blue}{73.0}& \textcolor{blue}{85.0} &90.8\\
			MLIP (Ours, ViT-B/16)& \textcolor{red}{57.0} & \textcolor{red}{89.0} & \textcolor{red}{89.4} & 	\textcolor{red}{90.0}& \textcolor{red}{53.0}&  \textcolor{red}{89.3} &  \textcolor{red}{90.0} &  \textcolor{red}{90.8}& \textcolor{red}{34.8} & \textcolor{red}{75.3}&\textcolor{red}{86.3}& \textcolor{red}{92.5}\\
			\bottomrule
	\end{tabular}}
	\caption{The image classification results in zero-shot scenarios and fine-tuning with 1\%, 10\%, and 100\% of the available training data are presented for the CheXpert, RSNA, and COVIDx datasets. The evaluation metric used is AUC [\%] for CheXpert and RSNA, and ACC [\%] for COVIDx. The best results achieved for each setting are highlighted in red, while the suboptimal results are highlighted in blue.}
	\label{Table 2}
\end{table*}

\subsection{Results}
\paragraph{Results on Medical Object Dection.}
We evaluate the ResNet-50-YOLOv3 architecture on the RSNA and Object CXR datasets. Our results, presented in Table \ref{od}, demonstrate a significant improvement over ConVIRT \cite{zhang2022contrastive}, GLoRIA \cite{huang2021gloria}, and MGCA \cite{wangmulti}. Notably, our method achieves superior performance using only 1\% of the data, surpassing alternative approaches that require 10\% or even 100\% of the data for fine-tuning.
\paragraph{Results on Medical Semantic Segmentation.}
In Table \ref{Table 3}, we present the semantic segmentation results (Dice [\%]) achieved on the SIIM and RSNA datasets using the ResNet-50-U-Net architecture. MLIP leverages contrastive learning and category-level approaches to achieve remarkable performance improvements, consistently obtaining the best results in various settings, as highlighted in red. Specifically, MLIP outperforms the state-of-the-art MGCA \cite{wangmulti} by 4.7\% on the RSNA dataset and 1.9\% on the SIIM dataset when fine-tuned with only 1\% of the training data.
\paragraph{Results on Medical Image Classification.}
Table \ref{Table 2} shows the medical linear classification results on RSNA and COVIDx datasets. We divide existing pre-trained methods into two categories: pre-trained on CheXpert \cite{irvin2019chexpert} and pre-trained on MIMIC-CXR\cite{johnson2019mimic}. The results of other approaches are from original papers, and we refer to \cite{wangmulti}, pre-train GLoRIA with MIMIC-CXR datasets. We evaluate these approaches in the zero-shot scenario and with 1\%, 10\% and 100\% of the data for fine-tuning, the results all outperform the SOTA. For a fair comparison, we pre-train our model with ResNet-50 and ViT-B/16 architecture. Except for the ViT-B/16 architecture, which yields comparable results to MGCA when fine-tuning is conducted using 100\% of the available data, all others achieve better performance than the same architecture.


\subsection{Ablation Study}
Table \ref{Table 4} presents ablation results on semantic segmentation for both RSNA and SIIM datasets. We observe that leveraging knowledge as an intermediate medium for aligning image-text pairs in contrastive learning substantially enhances the model's performance. Moreover, category-level contrastive learning aids in mitigating false negatives, thereby improving the model's generalization. Global contrastive learning acts as a performance lower bound, complementing local and category-level approaches and yielding promising outcomes. Table \ref{Table 5} presents the ablation results of our main contributions on the object detection task. Our proposed divergence encoder enhances feature diversity and enables the model to better adapt to challenging samples. With the assistance of expert knowledge, the alignment between medical images and medical reports becomes more efficient. Lastly, the proxy tasks designed in our approach strengthen the model's ability to discriminate negative samples.

\begin{table}
	\setlength{\belowdisplayskip}{0.0cm}
	\centering
	\resizebox{\linewidth}{!}{
		\begin{tabular}{cccccccccccc}
			\toprule
			\multicolumn{3}{c}{Tasks Setting} & \multicolumn{3}{c}{RSNA(Dice)} & \multicolumn{3}{c}{SIIM(Dice)}  \\
			Global ITA&Local ITA &Category-level ITA&1\% & 10\% & 100\% & 1\% & 10\% & 100\%
			\\ 
			\midrule
			&\checkmark& \checkmark&57.4&66.3&71.7&49.3&56.7&64.6\\
			\checkmark&&\checkmark&60.6&68.1&70.4&47.0&48.8&66.4\\
			\checkmark&\checkmark&&64.7&68.2&73.3&50.0&51.3&67.7\\
			\bottomrule
	\end{tabular}}
	\caption{Ablation study of our model on semantic segmentation task. It can observe that Global ITA has a more positive impact on the results, which can be attributed to the role of the divergence encoder.}
	\label{Table 4}
\end{table}

\begin{table}\tiny
	\setlength{\belowdisplayskip}{0.0cm}
	\centering
	\resizebox{\linewidth}{!}{
		\begin{tabular}{cccccccccccc}
			\toprule
			\multicolumn{3}{c}{Tasks Setting} & \multicolumn{3}{c}{RSNA(mAP)} & \multicolumn{3}{c}{Object CXR(mAP)}  \\
			DE&KA&TS+ITM&1\% & 10\% & 100\% & 1\% & 10\% & 100\%
			\\ 
			\midrule
			 \XSolidBrush&& &11.7&13.4&19.8&1.2&12.4&16.1\\
			& \XSolidBrush&&13.2&15.6&21.6&2.8&15.3&17.9\\
			&& \XSolidBrush&16.2&17.9&23.5&3.7&16.8&18.8\\
			\bottomrule
	\end{tabular}}
	\caption{Ablation study of our main contributions on object detection task, specially divergence encoder (DE), knowledge augmentation (KA) and text-swapping + image-text matching (TS+ITM).}
	\label{Table 5}
\end{table}

\subsection{Visualization}
To further understand the inner workings of MLIP, we present learned local correspondences between radiographs and medical reports in the form of heatmaps and showcase the performance of MLIP on downstream tasks (semantic segmentation and object detection) in the supplementary \ref{sec:Visualization}. The visual evidence supports that MLIP excels in fine-grained feature extraction, boosting accuracy.
\section{Conclusion}
We propose MLIP, a novel medical visual representation learning framework that integrates language information into the visual domain. Our model leverages multi-level image-text contrastive learning with divergence encoder and expert knowledge to enhance transfer capabilities for various downstream medical-related vision tasks. Our experimental results on multiple datasets demonstrate the effectiveness of MLIP, even in zero-shot scenarios and with limited annotated data.
{
    \small
    \bibliographystyle{ieeenat_fullname}
    \bibliography{main}

\begin{thebibliography}{64}
\providecommand{\natexlab}[1]{#1}
\providecommand{\url}[1]{\texttt{#1}}
\expandafter\ifx\csname urlstyle\endcsname\relax
  \providecommand{\doi}[1]{doi: #1}\else
  \providecommand{\doi}{doi: \begingroup \urlstyle{rm}\Url}\fi

\bibitem[Alsentzer et~al.(2019)Alsentzer, Murphy, Boag, Weng, Jin, Naumann,
  Redmond, and McDermott]{alsentzer2019publicly}
Emily Alsentzer, John~R Murphy, Willie Boag, Wei-Hung Weng, Di Jin, Tristan
  Naumann, WA Redmond, and Matthew~BA McDermott.
\newblock Publicly available clinical bert embeddings.
\newblock \emph{NAACL HLT 2019}, page~72, 2019.

\bibitem[Amizadeh et~al.(2020)Amizadeh, Palangi, Polozov, Huang, and
  Koishida]{amizadeh2020neuro}
Saeed Amizadeh, Hamid Palangi, Alex Polozov, Yichen Huang, and Kazuhito
  Koishida.
\newblock Neuro-symbolic visual reasoning: Disentangling.
\newblock In \emph{ICML}, pages 279--290. PMLR, 2020.

\bibitem[Ba et~al.(2016)Ba, Kiros, and Hinton]{ba2016layer}
Jimmy~Lei Ba, Jamie~Ryan Kiros, and Geoffrey~E Hinton.
\newblock Layer normalization.
\newblock \emph{arXiv preprint arXiv:1607.06450}, 2016.

\bibitem[Baumgartner et~al.(2021)Baumgartner, J{\"a}ger, Isensee, and
  Maier-Hein]{baumgartner2021nndetection}
Michael Baumgartner, Paul~F J{\"a}ger, Fabian Isensee, and Klaus~H Maier-Hein.
\newblock nndetection: a self-configuring method for medical object detection.
\newblock In \emph{MICCAI}, pages 530--539. Springer, 2021.

\bibitem[Ben-Younes et~al.(2017)Ben-Younes, Cadene, Cord, and
  Thome]{ben2017mutan}
Hedi Ben-Younes, Rmi Cadene, Matthieu Cord, and Nicolas Thome.
\newblock Mutan: Multimodal tucker fusion for visual question answering.
\newblock In \emph{ICCV}, pages 2612--2620, 2017.

\bibitem[Bodenreider(2004)]{bodenreider2004unified}
Olivier Bodenreider.
\newblock The unified medical language system (umls): integrating biomedical
  terminology.
\newblock \emph{NAR}, 32\penalty0 (suppl\_1):\penalty0 D267--D270, 2004.

\bibitem[Bordes et~al.(2013)Bordes, Usunier, Garcia-Duran, Weston, and
  Yakhnenko]{bordes2013translating}
Antoine Bordes, Nicolas Usunier, Alberto Garcia-Duran, Jason Weston, and Oksana
  Yakhnenko.
\newblock Translating embeddings for modeling multi-relational data.
\newblock \emph{Advances in neural information processing systems}, 26, 2013.

\bibitem[Caron et~al.(2020)Caron, Misra, Mairal, Goyal, Bojanowski, and
  Joulin]{caron2020unsupervised}
Mathilde Caron, Ishan Misra, Julien Mairal, Priya Goyal, Piotr Bojanowski, and
  Armand Joulin.
\newblock Unsupervised learning of visual features by contrasting cluster
  assignments.
\newblock \emph{NIPS}, 33:\penalty0 9912--9924, 2020.

\bibitem[Caron et~al.(2021)Caron, Touvron, Misra, J{\'e}gou, Mairal,
  Bojanowski, and Joulin]{caron2021emerging}
Mathilde Caron, Hugo Touvron, Ishan Misra, Herv{\'e} J{\'e}gou, Julien Mairal,
  Piotr Bojanowski, and Armand Joulin.
\newblock Emerging properties in self-supervised vision transformers.
\newblock In \emph{ICCV}, pages 9650--9660, 2021.

\bibitem[Chauhan et~al.(2020)Chauhan, Liao, Wells, Andreas, Wang, Berkowitz,
  Horng, Szolovits, and Golland]{chauhan2020joint}
Geeticka Chauhan, Ruizhi Liao, William Wells, Jacob Andreas, Xin Wang, Seth
  Berkowitz, Steven Horng, Peter Szolovits, and Polina Golland.
\newblock Joint modeling of chest radiographs and radiology reports for
  pulmonary edema assessment.
\newblock In \emph{MICCAI}, pages 529--539. Springer, 2020.

\bibitem[Chen et~al.(2021)Chen, Huang, Bisk, McDuff, and Gao]{chen2021kb}
Kezhen Chen, Qiuyuan Huang, Yonatan Bisk, Daniel McDuff, and Jianfeng Gao.
\newblock Kb-vlp: Knowledge based vision and language pretraining.
\newblock In \emph{ICML}, page 2021, 2021.

\bibitem[Chen et~al.(2020{\natexlab{a}})Chen, Kornblith, Norouzi, and
  Hinton]{chen2020simple}
Ting Chen, Simon Kornblith, Mohammad Norouzi, and Geoffrey Hinton.
\newblock A simple framework for contrastive learning of visual
  representations.
\newblock In \emph{ICML}, pages 1597--1607. PMLR, 2020{\natexlab{a}}.

\bibitem[Chen and He(2021)]{chen2021exploring}
Xinlei Chen and Kaiming He.
\newblock Exploring simple siamese representation learning.
\newblock In \emph{CVPR}, pages 15750--15758, 2021.

\bibitem[Chen et~al.(2020{\natexlab{b}})Chen, Li, Yu, El~Kholy, Ahmed, Gan,
  Cheng, and Liu]{chen2020uniter}
Yen-Chun Chen, Linjie Li, Licheng Yu, Ahmed El~Kholy, Faisal Ahmed, Zhe Gan, Yu
  Cheng, and Jingjing Liu.
\newblock Uniter: Universal image-text representation learning.
\newblock In \emph{ECCV}, pages 104--120. Springer, 2020{\natexlab{b}}.

\bibitem[Chen et~al.(2022)Chen, Li, and Wan]{chen2022align}
Zhihong Chen, Guanbin Li, and Xiang Wan.
\newblock Align, reason and learn: Enhancing medical vision-and-language
  pre-training with knowledge.
\newblock In \emph{ACM MM}, pages 5152--5161, 2022.

\bibitem[Cornia et~al.(2020)Cornia, Stefanini, Baraldi, and
  Cucchiara]{cornia2020meshed}
Marcella Cornia, Matteo Stefanini, Lorenzo Baraldi, and Rita Cucchiara.
\newblock Meshed-memory transformer for image captioning.
\newblock In \emph{CVPR}, pages 10578--10587, 2020.

\bibitem[Cui et~al.(2020)Cui, Zheng, and Wang]{cui2020unsupervised}
Wanyun Cui, Guangyu Zheng, and Wei Wang.
\newblock Unsupervised natural language inference via decoupled multimodal
  contrastive learning.
\newblock In \emph{EMNLP}, pages 5511--5520, 2020.

\bibitem[Cuturi(2013)]{cuturi2013sinkhorn}
Marco Cuturi.
\newblock Sinkhorn distances: Lightspeed computation of optimal transport.
\newblock \emph{NIPS}, 26, 2013.

\bibitem[De~Fauw et~al.(2018)De~Fauw, Ledsam, Romera-Paredes, Nikolov, Tomasev,
  Blackwell, Askham, Glorot, O’Donoghue, Visentin, et~al.]{de2018clinically}
Jeffrey De~Fauw, Joseph~R Ledsam, Bernardino Romera-Paredes, Stanislav Nikolov,
  Nenad Tomasev, Sam Blackwell, Harry Askham, Xavier Glorot, Brendan
  O’Donoghue, Daniel Visentin, et~al.
\newblock Clinically applicable deep learning for diagnosis and referral in
  retinal disease.
\newblock \emph{Nature medicine}, 24\penalty0 (9):\penalty0 1342--1350, 2018.

\bibitem[Dosovitskiy et~al.(2020)Dosovitskiy, Beyer, Kolesnikov, Weissenborn,
  Zhai, Unterthiner, Dehghani, Minderer, Heigold, Gelly,
  et~al.]{dosovitskiy2020image}
Alexey Dosovitskiy, Lucas Beyer, Alexander Kolesnikov, Dirk Weissenborn,
  Xiaohua Zhai, Thomas Unterthiner, Mostafa Dehghani, Matthias Minderer, Georg
  Heigold, Sylvain Gelly, et~al.
\newblock An image is worth 16x16 words: Transformers for image recognition at
  scale.
\newblock \emph{arXiv preprint arXiv:2010.11929}, 2020.

\bibitem[Engilberge et~al.(2018)Engilberge, Chevallier, P{\'e}rez, and
  Cord]{engilberge2018finding}
Martin Engilberge, Louis Chevallier, Patrick P{\'e}rez, and Matthieu Cord.
\newblock Finding beans in burgers: Deep semantic-visual embedding with
  localization.
\newblock In \emph{CVPR}, pages 3984--3993, 2018.

\bibitem[Esteva et~al.(2017)Esteva, Kuprel, Novoa, Ko, Swetter, Blau, and
  Thrun]{esteva2017dermatologist}
Andre Esteva, Brett Kuprel, Roberto~A Novoa, Justin Ko, Susan~M Swetter,
  Helen~M Blau, and Sebastian Thrun.
\newblock Dermatologist-level classification of skin cancer with deep neural
  networks.
\newblock \emph{nature}, 542\penalty0 (7639):\penalty0 115--118, 2017.

\bibitem[Faghri et~al.(2017)Faghri, Fleet, Kiros, and Fidler]{faghri2017vse++}
Fartash Faghri, David~J Fleet, Jamie~Ryan Kiros, and Sanja Fidler.
\newblock Vse++: Improving visual-semantic embeddings with hard negatives.
\newblock \emph{arXiv preprint arXiv:1707.05612}, 2017.

\bibitem[Farhadi and Redmon(2018)]{farhadi2018yolov3}
Ali Farhadi and Joseph Redmon.
\newblock Yolov3: An incremental improvement.
\newblock In \emph{CVPR}, pages 1--6. Springer Berlin/Heidelberg, Germany,
  2018.

\bibitem[He et~al.(2020{\natexlab{a}})He, Zhou, Xiao, Jiang, Liu, Yuan, and
  Xu]{he2020bert}
Bin He, Di Zhou, Jinghui Xiao, Xin Jiang, Qun Liu, Nicholas~Jing Yuan, and Tong
  Xu.
\newblock Bert-mk: Integrating graph contextualized knowledge into pre-trained
  language models.
\newblock In \emph{EMNLP}, pages 2281--2290, 2020{\natexlab{a}}.

\bibitem[He et~al.(2016)He, Zhang, Ren, and Sun]{he2016deep}
Kaiming He, Xiangyu Zhang, Shaoqing Ren, and Jian Sun.
\newblock Deep residual learning for image recognition.
\newblock In \emph{CVPR}, pages 770--778, 2016.

\bibitem[He et~al.(2020{\natexlab{b}})He, Fan, Wu, Xie, and
  Girshick]{he2020momentum}
Kaiming He, Haoqi Fan, Yuxin Wu, Saining Xie, and Ross Girshick.
\newblock Momentum contrast for unsupervised visual representation learning.
\newblock In \emph{CVPR}, pages 9729--9738, 2020{\natexlab{b}}.

\bibitem[He et~al.(2022)He, Chen, Xie, Li, Doll{\'a}r, and
  Girshick]{he2022masked}
Kaiming He, Xinlei Chen, Saining Xie, Yanghao Li, Piotr Doll{\'a}r, and Ross
  Girshick.
\newblock Masked autoencoders are scalable vision learners.
\newblock In \emph{CVPR}, pages 16000--16009, 2022.

\bibitem[Healthcare(2020)]{healthcare2020object}
J Healthcare.
\newblock Object-cxr-automatic detection of foreign objects on chest x-rays,
  2020.

\bibitem[Hjelm et~al.(2018)Hjelm, Fedorov, Lavoie-Marchildon, Grewal, Bachman,
  Trischler, and Bengio]{hjelm2018learning}
R~Devon Hjelm, Alex Fedorov, Samuel Lavoie-Marchildon, Karan Grewal, Phil
  Bachman, Adam Trischler, and Yoshua Bengio.
\newblock Learning deep representations by mutual information estimation and
  maximization.
\newblock \emph{arXiv preprint arXiv:1808.06670}, 2018.

\bibitem[Hsu et~al.(2018)Hsu, Weng, Boag, McDermott, and
  Szolovits]{hsu2018unsupervised}
Tzu-Ming~Harry Hsu, Wei-Hung Weng, Willie Boag, Matthew McDermott, and Peter
  Szolovits.
\newblock Unsupervised multimodal representation learning across medical images
  and reports.
\newblock \emph{arXiv e-prints}, pages arXiv--1811, 2018.

\bibitem[Huang et~al.(2021)Huang, Shen, Lungren, and Yeung]{huang2021gloria}
Shih-Cheng Huang, Liyue Shen, Matthew~P Lungren, and Serena Yeung.
\newblock Gloria: A multimodal global-local representation learning framework
  for label-efficient medical image recognition.
\newblock In \emph{ICCV}, pages 3942--3951, 2021.

\bibitem[Irvin et~al.(2019)Irvin, Rajpurkar, Ko, Yu, Ciurea-Ilcus, Chute,
  Marklund, Haghgoo, Ball, Shpanskaya, et~al.]{irvin2019chexpert}
Jeremy Irvin, Pranav Rajpurkar, Michael Ko, Yifan Yu, Silviana Ciurea-Ilcus,
  Chris Chute, Henrik Marklund, Behzad Haghgoo, Robyn Ball, Katie Shpanskaya,
  et~al.
\newblock Chexpert: A large chest radiograph dataset with uncertainty labels
  and expert comparison.
\newblock In \emph{AAAI}, pages 590--597, 2019.

\bibitem[Johnson et~al.(2019)Johnson, Pollard, Berkowitz, Greenbaum, Lungren,
  Deng, Mark, and Horng]{johnson2019mimic}
Alistair~EW Johnson, Tom~J Pollard, Seth~J Berkowitz, Nathaniel~R Greenbaum,
  Matthew~P Lungren, Chih-ying Deng, Roger~G Mark, and Steven Horng.
\newblock Mimic-cxr, a de-identified publicly available database of chest
  radiographs with free-text reports.
\newblock \emph{Scientific data}, 6\penalty0 (1):\penalty0 317, 2019.

\bibitem[Lee et~al.(2018)Lee, Chen, Hua, Hu, and He]{lee2018stacked}
Kuang-Huei Lee, Xi Chen, Gang Hua, Houdong Hu, and Xiaodong He.
\newblock Stacked cross attention for image-text matching.
\newblock In \emph{ECCV}, pages 201--216, 2018.

\bibitem[Li et~al.(2019)Li, Zhang, Li, Li, and Fu]{li2019visual}
Kunpeng Li, Yulun Zhang, Kai Li, Yuanyuan Li, and Yun Fu.
\newblock Visual semantic reasoning for image-text matching.
\newblock In \emph{ICCV}, pages 4654--4662, 2019.

\bibitem[Li et~al.(2023)Li, Laurence, Nie, Ren, and Deng]{li2023vipmm}
Zhe Li, T.~Yang Laurence, Xin Nie, BoCheng Ren, and Xianjun Deng.
\newblock Enhancing sentence representation with visually-supervised multimodal
  pre-training.
\newblock In \emph{ACM MM'23}, 2023.

\bibitem[Liu et~al.(2021)Liu, Wu, Ge, Fan, and Zou]{liu2021exploring}
Fenglin Liu, Xian Wu, Shen Ge, Wei Fan, and Yuexian Zou.
\newblock Exploring and distilling posterior and prior knowledge for radiology
  report generation.
\newblock In \emph{CVPR}, pages 13753--13762, 2021.

\bibitem[Liu et~al.(2020)Liu, Zhou, Zhao, Wang, Ju, Deng, and Wang]{liu2020k}
Weijie Liu, Peng Zhou, Zhe Zhao, Zhiruo Wang, Qi Ju, Haotang Deng, and Ping
  Wang.
\newblock K-bert: Enabling language representation with knowledge graph.
\newblock In \emph{AAAI}, pages 2901--2908, 2020.

\bibitem[Loshchilov and Hutter(2017)]{loshchilov2017decoupled}
Ilya Loshchilov and Frank Hutter.
\newblock Decoupled weight decay regularization.
\newblock \emph{arXiv preprint arXiv:1711.05101}, 2017.

\bibitem[Lu et~al.(2016)Lu, Yang, Batra, and Parikh]{lu2016hierarchical}
Jiasen Lu, Jianwei Yang, Dhruv Batra, and Devi Parikh.
\newblock Hierarchical question-image co-attention for visual question
  answering.
\newblock \emph{NIPS}, 29, 2016.

\bibitem[Qin et~al.(2022)Qin, Yi, Lao, and Li]{qin2022medical}
Ziyuan Qin, Huahui Yi, Qicheng Lao, and Kang Li.
\newblock Medical image understanding with pretrained vision language models: A
  comprehensive study.
\newblock \emph{arXiv preprint arXiv:2209.15517}, 2022.

\bibitem[Rajpurkar et~al.(2018)Rajpurkar, Irvin, Ball, Zhu, Yang, Mehta, Duan,
  Ding, Bagul, Langlotz, et~al.]{rajpurkar2018deep}
Pranav Rajpurkar, Jeremy Irvin, Robyn~L Ball, Kaylie Zhu, Brandon Yang, Hershel
  Mehta, Tony Duan, Daisy Ding, Aarti Bagul, Curtis~P Langlotz, et~al.
\newblock Deep learning for chest radiograph diagnosis: A retrospective
  comparison of the chexnext algorithm to practicing radiologists.
\newblock \emph{PLoS medicine}, 15\penalty0 (11):\penalty0 e1002686, 2018.

\bibitem[Redmon and Farhadi(2018)]{redmon2018yolov3}
Joseph Redmon and Ali Farhadi.
\newblock Yolov3: An incremental improvement.
\newblock \emph{arXiv preprint arXiv:1804.02767}, 2018.

\bibitem[Ronneberger et~al.(2015)Ronneberger, Fischer, and
  Brox]{ronneberger2015u}
Olaf Ronneberger, Philipp Fischer, and Thomas Brox.
\newblock U-net: Convolutional networks for biomedical image segmentation.
\newblock In \emph{MICCAI}, pages 234--241. Springer, 2015.

\bibitem[Russakovsky et~al.(2015)Russakovsky, Deng, Su, Krause, Satheesh, Ma,
  Huang, Karpathy, Khosla, Bernstein, et~al.]{russakovsky2015imagenet}
Olga Russakovsky, Jia Deng, Hao Su, Jonathan Krause, Sanjeev Satheesh, Sean Ma,
  Zhiheng Huang, Andrej Karpathy, Aditya Khosla, Michael Bernstein, et~al.
\newblock Imagenet large scale visual recognition challenge.
\newblock \emph{International journal of computer vision}, 115:\penalty0
  211--252, 2015.

\bibitem[Shih et~al.(2019)Shih, Wu, Halabi, Kohli, Prevedello, Cook, Sharma,
  Amorosa, Arteaga, Galperin-Aizenberg, et~al.]{shih2019augmenting}
George Shih, Carol~C Wu, Safwan~S Halabi, Marc~D Kohli, Luciano~M Prevedello,
  Tessa~S Cook, Arjun Sharma, Judith~K Amorosa, Veronica Arteaga, Maya
  Galperin-Aizenberg, et~al.
\newblock Augmenting the national institutes of health chest radiograph dataset
  with expert annotations of possible pneumonia.
\newblock \emph{Radiology: Artificial Intelligence}, 1\penalty0 (1):\penalty0
  e180041, 2019.

\bibitem[Tan and Bansal(2019)]{tan2019lxmert}
Hao Tan and Mohit Bansal.
\newblock Lxmert: Learning cross-modality encoder representations from
  transformers.
\newblock In \emph{EMNLP-IJCNLP}, pages 5100--5111, 2019.

\bibitem[Tian et~al.(2020)Tian, Krishnan, and Isola]{tian2020contrastive}
Yonglong Tian, Dilip Krishnan, and Phillip Isola.
\newblock Contrastive multiview coding.
\newblock In \emph{ECCV}, pages 776--794. Springer, 2020.

\bibitem[van~den Oord et~al.(2018)van~den Oord, Li, and
  Vinyals]{van2018representation}
Aaron van~den Oord, Yazhe Li, and Oriol Vinyals.
\newblock Representation learning with contrastive predictive coding.
\newblock \emph{arXiv e-prints}, pages arXiv--1807, 2018.

\bibitem[Vaswani et~al.(2017)Vaswani, Shazeer, Parmar, Uszkoreit, Jones, Gomez,
  Kaiser, and Polosukhin]{vaswani2017attention}
Ashish Vaswani, Noam Shazeer, Niki Parmar, Jakob Uszkoreit, Llion Jones,
  Aidan~N Gomez, {\L}ukasz Kaiser, and Illia Polosukhin.
\newblock Attention is all you need.
\newblock \emph{NIPS}, 30, 2017.

\bibitem[Velickovic et~al.(2017)Velickovic, Cucurull, Casanova, Romero, Lio,
  Bengio, et~al.]{velickovic2017graph}
Petar Velickovic, Guillem Cucurull, Arantxa Casanova, Adriana Romero, Pietro
  Lio, Yoshua Bengio, et~al.
\newblock Graph attention networks.
\newblock 1050\penalty0 (20):\penalty0 10--48550, 2017.

\bibitem[Wang et~al.()Wang, Zhou, Wang, Vardhanabhuti, and Yu]{wangmulti}
Fuying Wang, Yuyin Zhou, Shujun Wang, Varut Vardhanabhuti, and Lequan Yu.
\newblock Multi-granularity cross-modal alignment for generalized medical
  visual representation learning.
\newblock In \emph{NIPS}.

\bibitem[Wang et~al.(2020{\natexlab{a}})Wang, Lin, and Wong]{wang2020covid}
Linda Wang, Zhong~Qiu Lin, and Alexander Wong.
\newblock Covid-net: A tailored deep convolutional neural network design for
  detection of covid-19 cases from chest x-ray images.
\newblock \emph{Scientific reports}, 10\penalty0 (1):\penalty0 1--12,
  2020{\natexlab{a}}.

\bibitem[Wang et~al.(2021)Wang, Gao, Zhu, Zhang, Liu, Li, and
  Tang]{wang2021kepler}
Xiaozhi Wang, Tianyu Gao, Zhaocheng Zhu, Zhengyan Zhang, Zhiyuan Liu, Juanzi
  Li, and Jian Tang.
\newblock Kepler: A unified model for knowledge embedding and pre-trained
  language representation.
\newblock \emph{TACL}, 9:\penalty0 176--194, 2021.

\bibitem[Wang et~al.(2020{\natexlab{b}})Wang, Wang, and Zhu]{wang2020image}
Zhaobin Wang, E Wang, and Ying Zhu.
\newblock Image segmentation evaluation: a survey of methods.
\newblock \emph{Artificial Intelligence Review}, 53:\penalty0 5637--5674,
  2020{\natexlab{b}}.

\bibitem[Wei et~al.(2022)Wei, Fan, Xie, Wu, Yuille, and
  Feichtenhofer]{wei2022masked}
Chen Wei, Haoqi Fan, Saining Xie, Chao-Yuan Wu, Alan Yuille, and Christoph
  Feichtenhofer.
\newblock Masked feature prediction for self-supervised visual pre-training.
\newblock In \emph{CVPR}, pages 14668--14678, 2022.

\bibitem[Xu et~al.(2015)Xu, Ba, Kiros, Cho, Courville, Salakhudinov, Zemel, and
  Bengio]{xu2015show}
Kelvin Xu, Jimmy Ba, Ryan Kiros, Kyunghyun Cho, Aaron Courville, Ruslan
  Salakhudinov, Rich Zemel, and Yoshua Bengio.
\newblock Show, attend and tell: Neural image caption generation with visual
  attention.
\newblock In \emph{ICML}, pages 2048--2057. PMLR, 2015.

\bibitem[Yu et~al.(2021)Yu, Tang, Yin, Sun, Tian, Wu, and Wang]{yu2021ernie}
Fei Yu, Jiji Tang, Weichong Yin, Yu Sun, Hao Tian, Hua Wu, and Haifeng Wang.
\newblock Ernie-vil: Knowledge enhanced vision-language representations through
  scene graphs.
\newblock In \emph{AAAI}, pages 3208--3216, 2021.

\bibitem[Zawacki et~al.(2019)Zawacki, Wu, Shih, Elliott, Fomitchev, Hussain,
  Lakhani, Culliton, and Bao]{siim-acr-pneumothorax-segmentation}
Anna Zawacki, Carol Wu, George Shih, Julia Elliott, Mikhail Fomitchev, Mohannad
  Hussain, Paras Lakhani, Phil Culliton, and Shunxing Bao.
\newblock Siim-acr pneumothorax segmentation, 2019.

\bibitem[Zhang et~al.(2022)Zhang, Jiang, Miura, Manning, and
  Langlotz]{zhang2022contrastive}
Yuhao Zhang, Hang Jiang, Yasuhide Miura, Christopher~D Manning, and Curtis~P
  Langlotz.
\newblock Contrastive learning of medical visual representations from paired
  images and text.
\newblock In \emph{MLHC}, pages 2--25. PMLR, 2022.

\bibitem[Zhang et~al.(2017)Zhang, Chen, Sapkota, and Yang]{zhang2017tandemnet}
Zizhao Zhang, Pingjun Chen, Manish Sapkota, and Lin Yang.
\newblock Tandemnet: Distilling knowledge from medical images using diagnostic
  reports as optional semantic references.
\newblock In \emph{MICCAI}, pages 320--328. Springer, 2017.

\bibitem[Zhang et~al.(2019)Zhang, Han, Liu, Jiang, Sun, and
  Liu]{zhang2019ernie}
Zhengyan Zhang, Xu Han, Zhiyuan Liu, Xin Jiang, Maosong Sun, and Qun Liu.
\newblock Ernie: Enhanced language representation with informative entities.
\newblock In \emph{ACL}, pages 1441--1451, 2019.

\bibitem[Zheng et~al.(2021)Zheng, Lu, Zhao, Zhu, Luo, Wang, Fu, Feng, Xiang,
  Torr, et~al.]{zheng2021rethinking}
Sixiao Zheng, Jiachen Lu, Hengshuang Zhao, Xiatian Zhu, Zekun Luo, Yabiao Wang,
  Yanwei Fu, Jianfeng Feng, Tao Xiang, Philip~HS Torr, et~al.
\newblock Rethinking semantic segmentation from a sequence-to-sequence
  perspective with transformers.
\newblock In \emph{Proceedings of the IEEE/CVF conference on computer vision
  and pattern recognition}, pages 6881--6890, 2021.

\end{thebibliography}
}


\end{document}